# Additive Manufacturing Processes Protocol Prediction by Artificial Intelligence using X-ray Computed Tomography data

Sunita Khod[1], Akshay Dvivedi[2] and Mayank Goswami[1,#]

*[1]Divyadrishti Imaging Laboratory, Department of Physics,* Indian Institute of Technology Roorkee, Department of Mechanical Industrial Enggeinring
Roorkee, Uttarakhand-247667, India.

[#]mayank.goswami@ph.iitr.ac.in

## Abstract

The quality of the part fabricated from the Additive Manufacturing (AM) process depends upon the process parameters used, and therefore, optimization is required for apt quality. A methodology is proposed to set these parameters non-iteratively without human intervention. It utilizes Artificial Intelligence (AI) to fully automate the process, with the capability to self-train any apt AI model by further assimilating the training data.

This study includes three commercially available 3D printers for soft material printing based on the Material Extrusion (MEX) AM process. The samples are 3D printed for six different AM process parameters obtained by varying layer height and nozzle speed. The novelty part of the methodology is incorporating an AI-based image segmentation step in the decision-making stage that uses quality inspected training data from the Non-Destructive Testing (NDT) method. The performance of the trained AI model is compared with the two software tools based on the classical thresholding method. The AI-based Artificial Neural Network (ANN) model is trained from NDT-assessed and AI-segmented data to automate the selection of optimized process parameters.

The AI-based model is 99.3 % accurate, while the best available commercial classical image method is 83.44 % accurate. The best value of overall R for training ANN is 0.82. The MEX process gives a 22.06 % porosity error relative to the design. The NDT-data trained two AI models integrated into a series pipeline for optimal process parameters are proposed and verified by classical optimization and mechanical testing methods.

## 1. Introduction

The Conventional Manufacturing (CM) process of creating parts involves the casting, molding, forming, turning, and milling the materials [1]. Alternatively, Additive Manufacturing (AM) technology is evolving to fabricate complex-shaped specimens. Less material waste, less time consumption, the manufacturing of customized products, and the low cost of prototyping are the advantages of AM technology, which have gained considerable attention [2]. Additive Fabrication, Additive Process, 3D printing, and Additive Technique are synonyms for AM process [3]. The applications of AM technology include prototyping, aerospace, automotive, medical, energy, civil, robotics, and radiation shielding design [4].

ASTM International classified the AM process into seven categories. The categorization of processes according to ASTM ISO 52900 includes Vat Polymerization, Material Jetting, Binder Jetting, Material Extrusion, Powder Bed Fusion, Direct Energy Deposition, and Sheet Lamination. The process of AM can be understood as the interaction between the mass of the material and the energy required to form a single layer. The process of Powder Bed Fusion, Binder Jetting, Sheet Lamination, and Vat Polymerization falls in the AM process in which material mass is variable. Material Extrusion (MEX) and Material Jetting (MJT) are grouped as variable energy processes. In Direct Energy Deposition, material mass and energy are variable while forming a single layer [5].

### 1.1. Process Control

Additive Manufacturing still lags behind subtractive and generative manufacturing techniques, as far as technical maturity is concerned, in many aspects, e.g., in process control and material selection [6]. The quality of the part fabricated from either of the AM processes may differ from the Computer-Aided Design (CAD) model in terms of geometric dimensions. Part manufactured by either of the methods may contain defects (porosity and different surface roughness) arising due to either choice of inapt fabrication process and/or sub-optimal process parameters.

The performance and quality of the part fabricated thus depend upon the process parameters of the corresponding AM process[7], [8]. These parameters are mostly decided by an operator. The quality of printing may vary from operator to operator. Quality inspection is thus necessary to achieve the desired quality, especially for high-value and/or critical components used for aerospace and medical applications [9], [10].

## 1.2.    Process Control Optimization

Quality assurance improvement can be achieved by choosing the appropriate AM method and optimizing the process parameters for the AM method used. The simplest methodology is to vary the combination of sub-optimal process type and process parameters, including the material of choice, and assess the quality iteratively until satisfactory quality is achieved. It is a costly and time-consuming approach. However, an operator may gain enough experience to reduce these iterations after a certain operational period. The accuracy and speed of this methodology also depend on the accuracy of the assessment technique; otherwise, the operator would gain relatively incorrect experience. The optimization of process parameters for AM processes can utilize analytical Destructive testing (DT) and/or functional Non-Destructive Technique (NDT) methods. The X-ray Computed Tomography (X-ray CT) technique is categorized under the NDT method. The correlation between DT(Tensile Test) and NDT (X-ray and Ultrasonic) data of AM samples for material extrusion and jetting processes are reported in the literature. The correlation is found to be linear[11], [12].

The optimality criteria can be pre-decided physical parameters such as the amount of (a) porosity, (b) surface roughness, and/or (c) fiber alignment. The decision-making step may involve classical optimization techniques or Artificial Intelligence (AI). The classical optimization method optimizes the value of optimal criteria by minimizing printing errors between the CAD model and QA data from the printed sample. It may establish an analytical relation between process parameters and the optimality criteria by a curve fitting the iteratively obtained experimental data that can be used later [13].

## 1.3.    Utility of AI for Quality Assessment (QA)

The quality-assessed AM process has three stages where a person is involved: (a) to decide the printers' process parameters, (b) to assess the quality of print for the first decision, and (c) iteratively change the decision 'a' if 'b' results in unsatisfactory performance. With experience, one does not need a 'c'. AI can be used (a) to augment and preserve the manual experience, (b) to avoid any manual error, (c) to fasten the AM process, and (d) to save the environment by printing non-iteratively. QA of AM can be done by tracking the optimality parameters using AI.

## 1.4.    QA and AI

Functional NDT may require multi-material image segmentation. Several signal, data, or image-processing AI models are reported in the literature. Some of these may not be versatile enough to handle particular field-specific data. However, their flexible architecture can be tuned to handle data irrespective of the field. A typical Machine Learning (ML) model works well for data having a good binomial distribution fit. ML models are known to perform well for inadequate and poor-quality training data [14]. Deep learning-based convolutional neural networks, such as U-net, Residual neural network, and several other variant models, have been shown to work well for feature extraction and image segmentation. U-net is most widely used for image segmentation because it is faster and provides detailed segmentation for fine features [15], [16], [17]. These data-hungry models' performance improves with the amount of training data with diverse characteristics [18]. Guan Lin Chen et al. tested machine-learning methods to detect misalignment in the intended fiber directions in 3D-printed Fiber Reinforcement Composites using X-ray CT  [19]. However, the AM process parameters correction step for reducing this misalignment is not discussed. Their work has used U-net for quantifying binarized statistical image features to assess the quality of the extrusion AM process, and only a very small 36 $mm^2$ single sample is scanned using a commercial X-ray CT. Porosity analysis of X-ray CT data of four 3D printed samples using U-net is done by Vivian Wen Hui Wang et al. [20].

### 1.4.1.    Process parameters optimization using AI

The AI model can be trained with DT or NDT data for AM process parameters optimization. So, depending on the data and method used for AM process parameters optimization, the process can be classified: (a) destructive technique assessed classically optimization driven, (b) destructive technique assessed AI-driven, (c) NDT assessed classically optimization driven and (d) NDT assessed AI-driven. NDT methods are considered relatively accurate, cost-effective, and faster than destructive methods for quantitative analysis [21], [22].

The DT assessed classically optimization method uses tensile and compression tests to optimize process parameters by assessing parameters that include Load, extension, breaking point, tensile strength, Young's modulus, yield strength, etc. [23], [24], [25] [26], [27]. DT-assessed AI-driven method uses tensile strength and compression tests to train the model and predict the optimized process parameters [28], [29], [30], [31]. The NDT assessed classically optimization method utilizes either porosity, surface roughness, fiber orientation, pore size, or etc., data obtained from NDT (X-ray CT/ultrasound) to minimize defect and optimize process parameters [32], [33], [34].

The optimization of AM process parameters by using DT data to train the ANN model for automation of AM process parameter selection is reported in the literature. The ANN model is trained for a large data set and then predicted for the test data set. The output variable of the ANN model is *minimized/maximized* for the given set of input variables to predict the optimized process parameters. More information about output/input is explained in section 2.2.7. Feed Forward, Genetic Algorithms, Multilayer Perceptron, linear regression, feed-forward backpropagation, etc. are some of the preferred ANN models. Several algorithms are developed to optimize ANN model parameters. Particle Swarm Optimization, Artificial Bee Colony, Backtracking search algorithm, Gradient descent, etc., are the most used algorithm techniques [35].

## 1.5. Motivation

The decision-making step to choose the apt AM process and respective process parameters needs to be automated. A feedback loop may be incorporated using quality inspection NDT to minimize printing errors. *An AI-driven process control methodology incorporating NDT data in training is not reported.*

This study mixes these missing connections by presenting an algorithm to predict optimal process parameters for the AM process using NDT data-trained AI models. Two AI models are sequentially integrated into a series pipeline: (a). AI model 1 for segmentation of NDT data and (b). AI model 2 for process protocol prediction. The primary goal is to automate the AM process and improve the print quality.

# 2. Methodology

## 2.1. Investigation steps

The motivation of this study is to optimize AM process parameters using two AI models trained with NDT data. A flowchart of the proposed methodology is shown in Fig. 1. An AI model 1 (U-net) is trained to identify pores in X-ray CT images. The AI model segments the X-ray CT data and provides a porosity estimate. The estimated porosity and corresponding AM process parameters are then used to train AI model 2 (ANN). This AI model, when fed with a combination of AM process parameters, predicts the possible porosity in the sample.

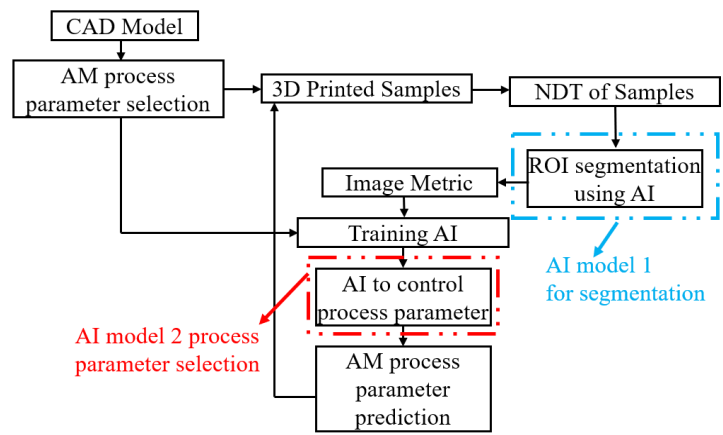

**Fig. 1:** Flowchart of the proposed methodology for process parameters selection

Estimating surface roughness by classical methods and mechanical testing can also be used to optimize process parameters. So, the quantitative surface roughness analysis and compression test are performed to find the optimized process parameters and to verify the results obtained from the proposed methodology. The contribution of these results is to validate the results of AI models trained with NDT data.

The methodology of this paper is divided into four steps, as shown in Fig. 2. The first two steps consist of the implementation of two AI models for the optimization of the AM process parameters. The third and fourth steps validate results obtained from the two AI models implemented in the first two steps.

The first step is modeling two designs, Design 1 (D1) and Design 2 (D2) of different dimensions and inner profiles using the CAD software Solidworks 2019. The CAD files are converted into Standard Triangle or Tessellation Language (STL) files. A gcode file is created to 3D print the samples. The samples are 3D printed with three printers based on the MEX AM process that allows varying process parameters. The 3D-printed samples are scanned using

X-ray $\mu$CT for NDT data collection. The X-ray CT data is segmented using AI model 1 to quantify defects and porosity. Then, the porosity is estimated from AI model 1 and two software based on the classical thresholding method. An optimal tool for porosity estimation is found.

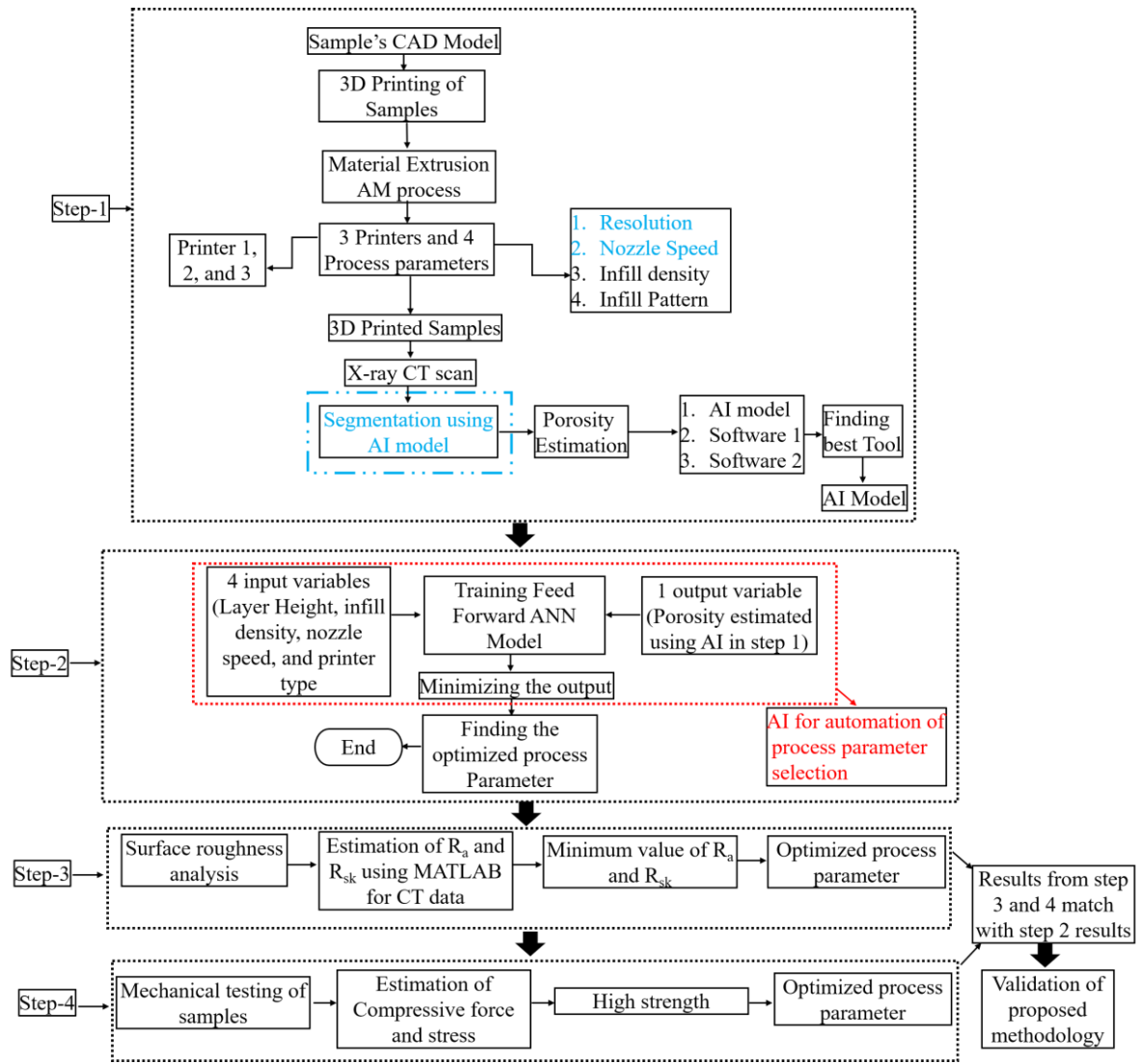

**Fig. 2:** Flowchart for the implementation of the methodology adopted in this study.

The second step is the implementation of an AI model 2 for process parameters control. The AI model 2 is trained with four input variable variables (layer height, nozzle speed, infill density, and printer type) and one output variable, which is porosity estimated from the AI model in the first step. The model is tested for different input parameters, and the corresponding porosity value is predicted. [R2 C2b:] The process parameters giving the voids in the printed sample design near to the designed ones and minimum defects due to printing is considered as the optimal process parameters.

The third step is the quantitative surface roughness analysis. The quantitative analysis estimates average roughness, $R_a$, and skewness parameters, $R_{sk}$, for X-ray CT data using MATLAB code.

The compression test of the 3D-printed sample is performed in the fourth step. The compressive force and stress are estimated as a function of porosity.

## 2.2.    Materials and Methods

### 2.2.1.    CAD Modeling and 3D Printing Process

The dimensions of the CAD for designs D1 and D2 are 8 (L) $\times$ 8 (W) $\times$ 26 (H) mm and 6 (L) $\times$ 6 (W) $\times$ 17.50 (H) mm, respectively, as shown in Fig. 3(a) and 3(c). The two CADs are also modeled with the designed distribution of voids. The minimum void size in CAD is chosen according to the spatial resolution of X-ray and printable resolution of 3D printers [36], [37]. D1 contains cuboids (minimum size 0.20 (L) $\times$ 0.20 (W) $\times$ 0.20 (H) mm), and D2 has a mixed distribution of cuboids (minimum size $0.20 \times 0.20 \times 0.20$ mm) and spherical cavities (minimum size 0.20mm (Diameter)) as shown in Fig. 3(b) and 3(d). Fig. SF1 of supplementary section S1 provides detailed information about the pattern and sizes of the inserted voids. Design 1 and Design 2 have ~1.45 % and 2.48% porosity (by design), respectively. A detailed description of the 3D printing process is given in section S1 of supplementary data.

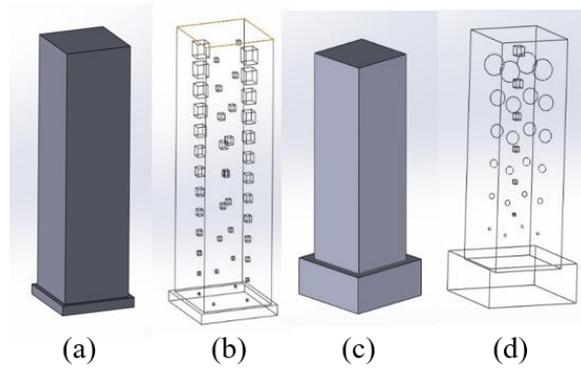

(a)                (b)                (c)                (d)

**Fig. 3** CAD and inner profile of Design 1 (D1) and Design (D2). (a). CAD image of D1; (b). Void distribution in D1; (c). CAD image of D2; (d). Void distribution in D2.

### 2.2.2. Printer Type, Material, and Process Parameters

The three 3D printers utilized in this study are listed in Table 1. This is because of 3 different manufacturers and biasedness towards the manufacturer. The resolution for each 3D printer is different, as listed in Table 1. The material used for printing the samples by the three printers is Poly-lactic Acid (PLA). The melting point of PLA varies from 195°C-220°C. PLA is a biodegradable thermoplastic polyester obtained from sugarcane or corn starch.

**Table 1:** List of Printers with the printing process and Materials

| S. No. | Printer Name | Resolution (μm) | Printing Process | Material |
|--------|-------------|-----------------|------------------|----------|
| 1. | Ultimaker Extended 2++ | 20-200 | MEX | PLA |
| 2. | Delta Wasp | 50 | MEX | PLA |
| 3. | Raise E2 | 50 | MEX | PLA |

An experiment is designed to print the samples by varying different process parameters. The varied process parameters include layer height and nozzle speed. The other parameters can also be varied. However, the motivation of the study is to propose an AI-based methodology for optimizing process parameters. So, if we include other parameters and vary that other parameters or increase the number of process parameters, the methodology for the process will remain the same, the results for the optimal process parameters may vary depending on the inclusiveness of the process parameters type.

The layer height varies from 50μm to 70μm while the nozzle speed is 30mm/s and 35mm/s. The infill density is 100%, and the infill pattern type is Grid. The Nozzle Temperature is maintained at 220°C, and the Nozzle diameter is 0.4mm. The bed temperature of the printer is kept at 45° C. The list of process parameters is given in Table 2.

**Table 2:** List of Process Parameters used for printers 1, 2, and 3 listed in Table 1

| S. No | Layer Height (μm) | Nozzle Speed (mm/s) | Infill Density (%) | Infill Pattern |
|-------|-------------------|---------------------|--------------------|----------------|
| 1. | 50 | 30 | 100 | Grid |
| 2. | 55 | 30 | 100 | Grid |
| 3. | 60 | 30 | 100 | Grid |
| 4. | 65 | 30 | 100 | Grid |
| 5. | 70 | 30 | 100 | Grid |
| 6. | 50 | 35 | 100 | Grid |

### 2.2.3. 3D Printed Samples

In total, 36 samples are 3D printed. This number is due to 6 process parameters (Table 2) for 2 Designs (D1 and D2) of different dimensions by three printers based on the MEX AM process (listed in Table 1). The samples printed from designs D1 and D2 are labeled as $Sample_{D1}$ and $Sample_{D2}$ . The time taken (2 hrs. 30 mins) for printing the samples is approximately the same for all the 3D printers. The samples with the black texture are printed from printer 1, with a blue color from printer 2, and white from printer 3, as shown in Fig. 4.

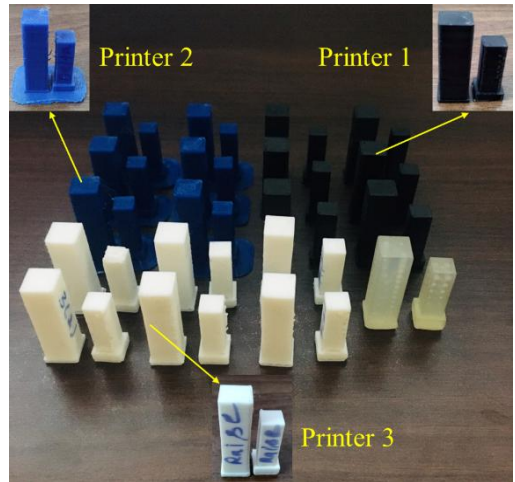

**Fig. 4** 3D printed samples

### 2.2.4. X-Ray Computed Tomography

A 35 kV, 1 mA setting X-ray CT system with a 1K x 1K pixel resolution detector is used for scanning and NDT data collection. 360 scans/rotations, each with 0.5 sec exposure time, are performed. The X-ray CT system used to scan the printed samples is shown in  Fig. 5.

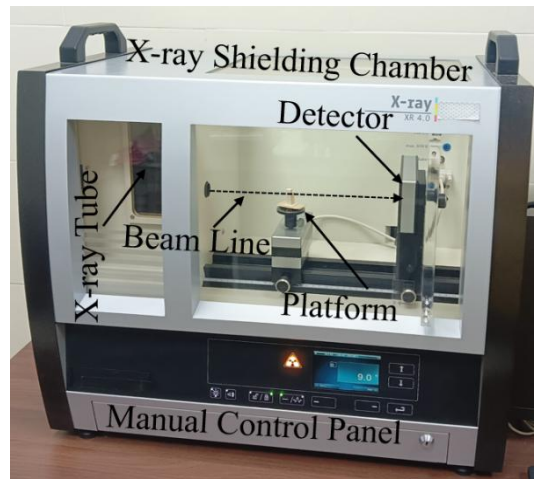

**Fig. 5** X-ray Computed Tomography system used to scan the printed samples.

### 2.2.5. Training of AI model using NDT data for QA

An AI model (referred to as AI model 1) is used to segment the collected X-ray CT data. The AI model used in this study is named U-net, which is a U-like symmetrical structure model consisting of an encoder (contracting) and a decoder (expansion) path, as shown in Fig. 6. Each down-sampling step in the contracting path consists of two successive 3×3 Convolutions followed by a Rectified Linear Unit (ReLU) and a 2×2 Max Pooling operation with stride 2. The one down-sampling step doubles the feature channels while reducing the resolution by half. The structure of the expansion path is similar to the contraction path, except the down-sampling step is replaced by the up-sampling step. Each up-sampling step in the expansion path consists of three operations. The first operation is a 2×2 up-convolution that halves the number of feature channels in the image. The second operation is the concatenation with the cropped feature map of the corresponding layer of the contracting path. This operation retrieves the spatial information lost during pooling operations. The third operation is the two successive 3×3 convolution operations, each followed by a ReLU. Both the down-sampling and up-sampling steps are repeated four times. The final results are generated by applying a 1×1 convolution at the final layer. The network architecture has four deep layers with 23 convolutional layers, as shown in Fig. 6 [15], [16], [17].

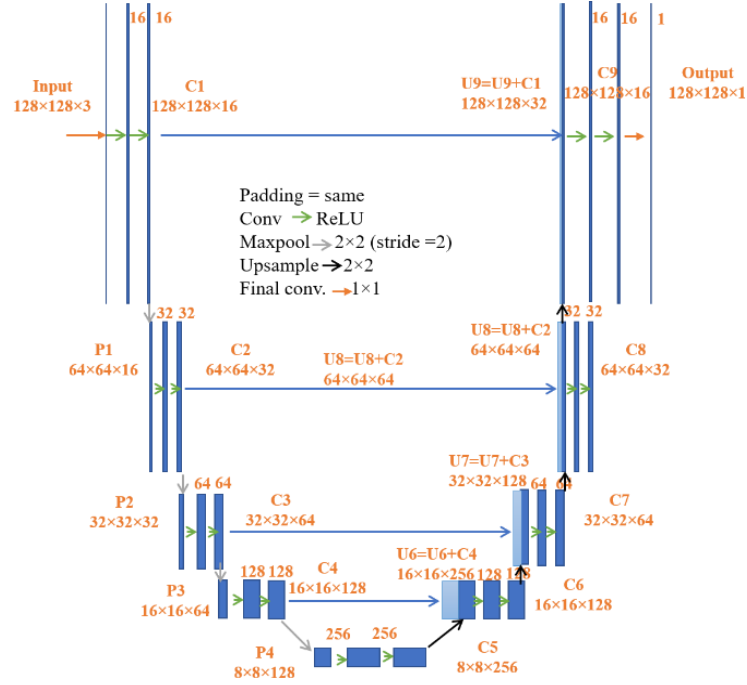

**Fig. 6** Structure of the AI model used for Image segmentation.

The RGB X-ray CT images and the corresponding mask data are used to train and predict the model. The input is a 3-channel image with [128, 128] size, while the mask is a 1-channel image with [128, 128] size. The predicted output is also a 1-channel image of size [128, 128]. Ninety percent of data is used for training, while 10 percent is used for validation. The model is trained with a batch size of 10 and 50 epochs, and corresponding loss and accuracy are measured and plotted. A threshold of 0.8 is applied to predict the final segmented labels. A total of 1867345 parameters are estimated using this model, out of which 1867345 are trainable while the number of non-trainable parameters is zero.

## 2.2.6. Porosity Estimation

Porosity measures the void spaces present in a sample or specimen. Quantitatively, it is defined as the ratio of pore volume to the total volume of the bulk or sample [38]. The porosity formula is given by: -

$$\Phi = \frac{V_{pore}}{V_{bulk}} \times 100 \qquad (1)$$

Where $V_{pore}$ and $V_{bulk}$ are the total volume of pores in the sample and the total volume of the sample, respectively. Porosity is a significant characteristic of a material or sample that defines the sample's strength, performance, and operational life. The presence of porosity in a sample affects the mechanical properties of the model as well. High percentages of porosity may render the piece useless. It may also lead to the formation of cracks. Sometimes, the point of failure coincides with the location of the pores. Therefore, porosity analysis helps in analyzing the quality of the sample.

The porosity for all the 3D printed samples is estimated by using the X-ray CT data. The two software viz software 1 and 2 based on thresholding method as listed in Table 3 and one AI tool (U-net) are used for porosity estimation. The porosity estimated by the AI model 1 is used to train the AI model 2 for process parameters optimization.

**Table 3:** List of software tools based on the thresholding method for porosity analysis

| S.No. | Referred name in text | Software Name |
|-------|----------------------|---------------|
| 1.    | Software 1            | Dragonfly     |
| 2.    | Software 2            | VG Studio Max |

## 2.2.7. Training AI model for process control

The AI model (referred to as AI model 2) used for process control is the Artificial Neural Network (ANN) model. ANN is a supervised ML method that predicts the outcome of a given input. The algorithm utilized in training the model is Gradient descent with momentum and adaptive Learning rate. It is a specific variant of the gradient descent optimization algorithm used in training machine learning models. The inbuilt NN tool of MATLAB and corresponding functions available in the module are used for training and validation of the model [39]. Studies using the algorithms for destructive testing methods are reported [30] [40]. The same model can also be implemented using Opensource libraries of python/C++ [41], [42].

We tested the model with three structures viz 4-4-1, 4-8-1, and 4-10-1 to select the optimal network structure. The neural network with a 4-4-1 structure and a high correlation coefficient is considered the optimal network structure. This choice solely is based upon the assumption (and can be modified by others) that selection or prediction of process parameter is simple task where overfitting needs to be avoided as the data is limited. The structure of the ANN used is 4-4-1, that is four neurons in the input layer, four neurons in the hidden layer, and 1 neuron in the output layer. The four neurons in the input layer corresponds to the four-input variable viz layer height, nozzle speed, infill density, and printer type. The four neurons in the hidden layer are choose according to the high value of correlation coefficient for these number of neurons. One neuron in the output layer corresponds to one output variable viz porosity. The porosity (NDT data) of all the samples estimated by AI model (section 2.6) is used as the output variable in training the model. 70 % sample data is used for training, 15 % for testing, and 15 % for validation.

The input layer consisting of four neurons with each neuron representing one input variable is represented as:

$$x = [x_1, x_2, x_3, x_4] \tag{2}$$

Where $x_1, x_2, x_3,$ and $x_4$ are layer height, nozzle speed, infill density, and printer type.

The neurons in the hidden layer are connected to each neuron in the input layer by weighted connections, $w_{ij}$. The weight of each neuron identifies the contribution of the input variable. The weighted input to $j^{th}$ neuron of the hidden layer is expressed as:

$$z_j = \sum_{i=1}^{4} w_{ij} x_i + b_j \tag{3}$$

Where $z_j$ is the input to $j^{th}$ neuron in the hidden layer, $w_{ij}$ is the weight of connection for $i^{th}$ input neuron to the $j^{th}$ hidden neuron, $x_i$ is the $i^{th}$ input variable, and $b_j$ is the bias term for $j^{th}$ hidden neuron.

The weighted input ($z_j$) is passed through the $j^{th}$ hidden neuron in the hidden layer by an activation function, $\psi_h$ (sigmoid, ReLU, tanh, etc.). We have used log-sigmoid activation function for hidden layer of the trained ANN model. The output of the $j^{th}$ hidden neuron is given by:

$$h_j = \psi_h(z_j) = \psi_h\left(\sum_{i=1}^{4} w_{ij} x_i + b_j\right) \tag{4}$$

The output of the hidden layer is represented as:

$$h = [h_1, h_2, h_3, h_4] \tag{5}$$

The output layer consists of one-neuron corresponding to porosity and computes the porosity. The weighted input to the output neuron, $z_o$ is expressed as:

$$z_o = \sum_{j=1}^{4} w_j^o h_j + b^o \tag{6}$$

Where $w_j^o$ is the weight connecting the $j^{th}$ hidden neuron to the output neuron, $h_j$ is the output of the $j^{th}$ hidden neuron, and $b^o$ is the bias term for the output neuron.

The weighted input ($z_o$) is passed through the output neuron by the activation function, $\psi_o$. The activation function can be sigmoid, linear, identity, or any other function. We have used tan-sigmoid activation function for output layer of the trained ANN model. The final output or predicted porosity ($\phi_{pred}$) of the trained ANN is given by:

$$\phi_{pred} = \psi_o \left( \sum_{j=1}^{4} w_j^o \psi_h \left( \sum_{i=1}^{4} w_{ij} x_i + b_j \right) + b^o \right) \tag{7}$$

The weights ($w_{ij}$, $w_j^o$) and biases ($b_j$, $b^o$) are calculated during training the ANN with data containing a set of process parameters as input variables and corresponding porosity values as output variable. These weights and biases are optimized using algorithms such as Gradient descent during the training of the model. The loss function, which is Mean Square Error (MSE) for regression is also calculated during the training of the ANN and is given by:

$$Loss = \frac{1}{N} \sum_{i=1}^{N} (\phi_i - \phi_{pred,i})^2 \tag{8}$$

Where N is the number of training samples, and $\phi_i$ is the original value of porosity, and $\phi_{pred,i}$ is the predicted value of porosity.

Fig. 7 shows the architecture of the AI model 2 and Fig. 8 shows the structure of the Neural Network.

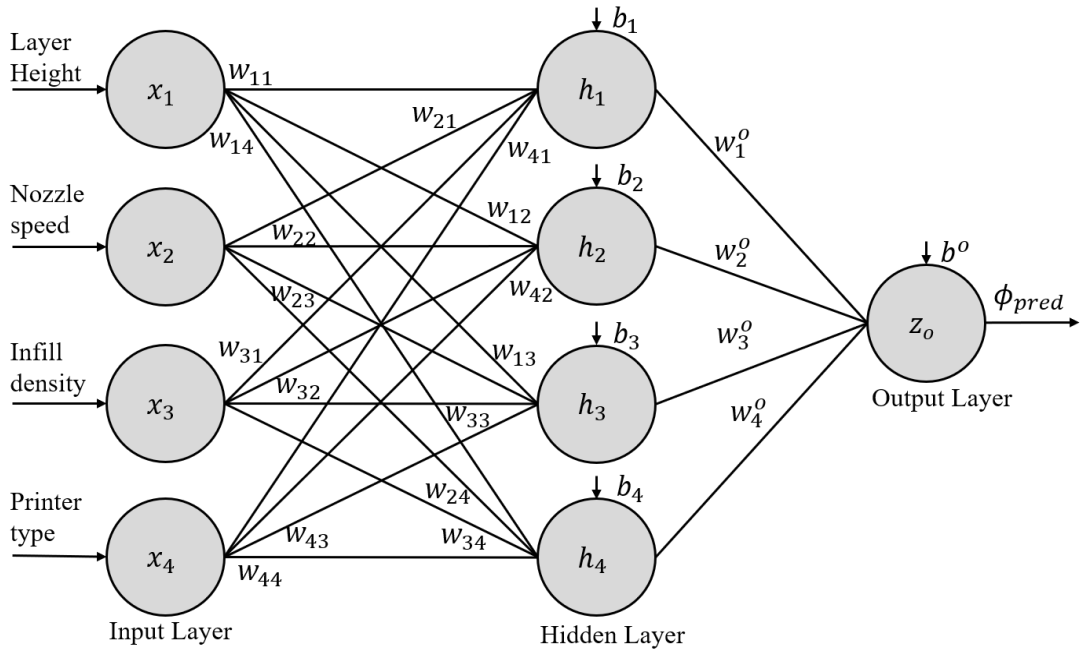

**Fig. 7** Architecture of the AI model 2 with one input layer, one hidden layer, and one output layer.

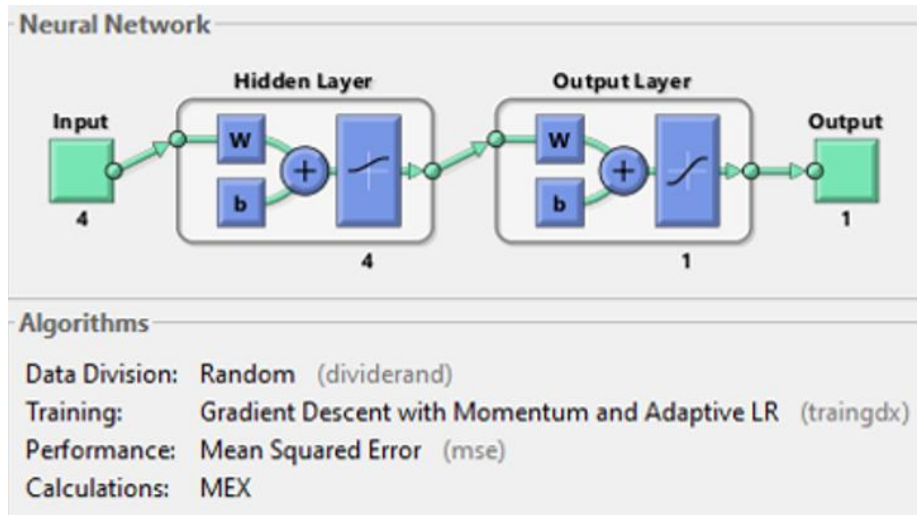

**Fig. 8** Structure of the Neural Network.

### 2.2.8. Surface Roughness Analysis

The measurement of surface roughness is another parameter used to evaluate the quality of the sample. Roughness is a measurement of a component's surface finish or texture. Quantitatively, it is defined by the deviations in the direction of the normal vector of a real surface from its ideal direction. Based on the necessity of the application, a surface can be classified as rough or smooth depending on how large or small the deviation is from the ideal flat surface. Roughness can be used as one of the predictors of the life of a component since irregularities on the surface may form nucleation sites for cracks.

The arithmetic average roughness ($R_a$) and Skewness parameters ($R_{sk}$) can be determined to determine the surface roughness. $R_a$ is the arithmetic average of profile height from the mean line. The Skewness parameters $R_{sk}$ measure the asymmetry of the profile height about the mean line. $R_a$ is measured in the units of µm while $R_{sk}$ is a dimensionless quantity [43]. Mathematically, $R_a$ and $R_{sk}$ are given by:-

$$R_a = \frac{1}{l_r} \int_0^{l_r} |Z(x)| dx \qquad (9)$$

$$R_{sk} = \frac{1}{R_q{}^3} \left[ \frac{1}{l_r} \int_0^{l_r} Z(x)^3 dx \right] \qquad (10)$$

Where $l_r$ is the sampling length, $R_q$ is the root mean square average of profile height, and $Z(x)$ is the ordinate distance values within the sampling length [43]. The parameters $R_a$ and $R_{sk}$ are estimated using MATLAB for the NDT scan data (X-ray CT data).

[R2 C3a:] According to classical method of process parameters optimization, the optimal process parameters correspond to the set of parameters that prints the sample with minimum surface roughness that is value of $R_a$ and $R_{sk}$ parameter is minimum.

### 2.2.9. Mechanical Testing of Samples

The compression test of all the 3D-printed samples ($Sample_{D1}$ and $Sample_{D2}$) is performed using the Instron 5982 tensile testing machine, as shown in Fig. 9. All the samples are placed in the same position during the compression test. The starting point height of the test is also fixed for all the samples. The test is performed with a strain rate of 2mm/min, maximum Load of 20kN, and maximum extension of 15mm. The testing process is shown in the movie link attached here. (Link)

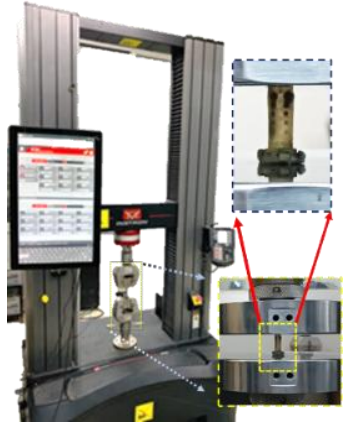

**Fig. 9** Compression Test Setup

## 3. Experimental Results

### 3.1. Training and Testing of U-net

The X-ray CT data generated from scanning the samples is used for training and testing the 28-convolutional layered AI model 1. The annotated data corresponding to X-ray CT data is created using MATLAB and used as mask data to train the AI model. The method used to create the mask data is the threshold method using the im2bw function. The model is trained and tested for different samples separately. The labels predicted using the AI model 1 for $Sample_{D1}$ and $Sample_{D2}$ printed from Printer 1 for the first process parameters (50µm layer height, 30mm/s nozzle speed, 100% infill density) is shown in Fig. 10. The labels for other samples are shown in Fig. SF2 of supplementary section S2. The blue arrows shown in Fig. 10 correspond to the printer error, the red arrows represent the designed voids, and the yellow arrows show the background segmented using the AI model 1.

Original Image          Segmented Image

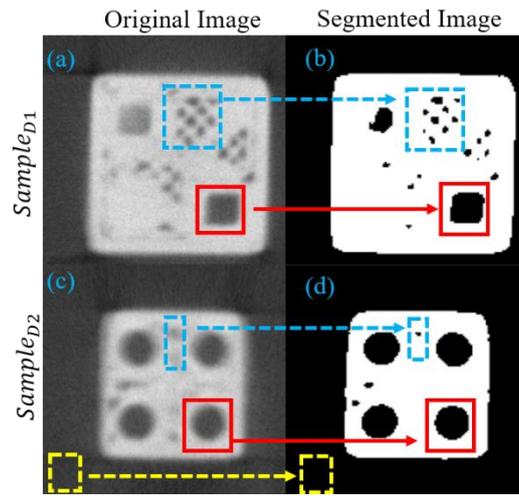

**Fig. 10** Segmented X-ray CT image using AI model 1

The parameters calculated for training and validation of the AI model 1 include loss, accuracy, validation loss, and validation accuracy and plot for $Sample_{D1}$ is shown in Fig. 11. For $Sample_{D2}$ the analysis is shown in Fig SF3 of supplementary section S2. All these parameters are calculated for 50 epochs. An average loss of 3 % and accuracy of 98 % is observed in the training data for X-ray CT images of samples printed from the MEX AM process. The validation loss and accuracy follow the same trend; however, small fluctuations are observed due to model prediction deviations. The variations in the loss and accuracy plot of all three 3D printers is small. However, the highest validation accuracy is observed for Printer 2, followed by Printer 3 and 1, respectively. The higher accuracy of the AI model shows the accuracy of the model for label prediction. The predicted labels are further used for porosity estimation in the next section. The same observations hold for the loss and accuracy of $Sample_{D2}$ data.

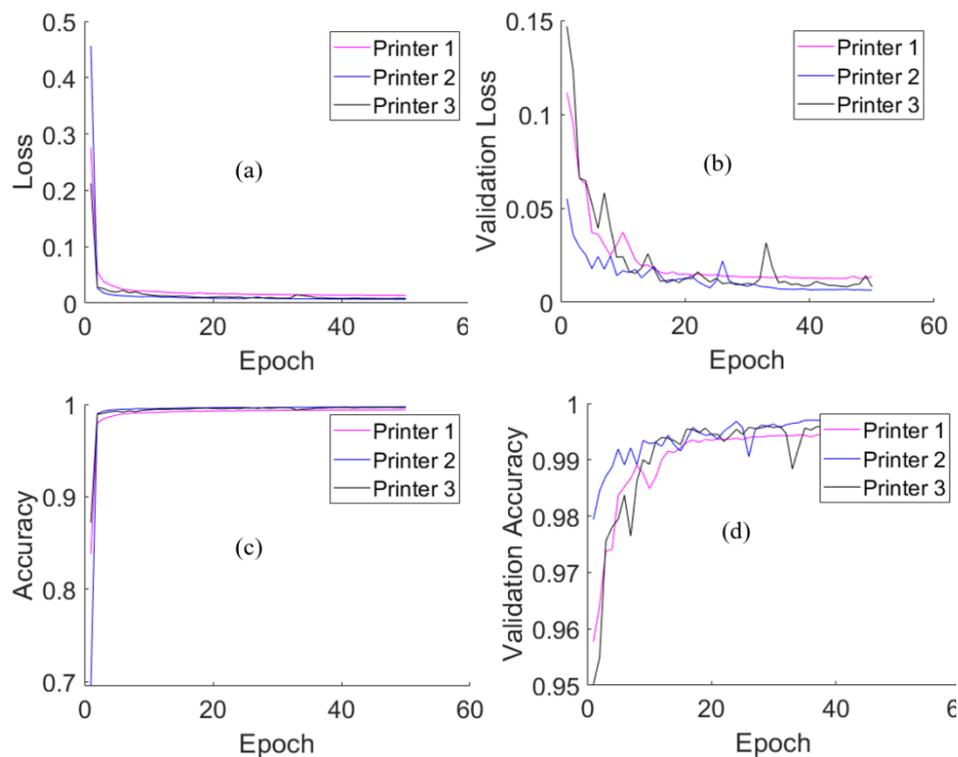

**Fig. 11** Training and Testing of $Sample_{D1}$ CT data **(a).** Loss **(b).** Validation Loss **(c).** Accuracy **(d).** Validation Accuracy

## 3.2. Porosity Analysis and Optimal Computational Tool

The porosity and designed void volume in the 3D printed samples are estimated using Software 1 and Software 2, which are based on the thresholding method. The analysis is also done for the X-ray CT data segmented using an AI model to compare the optimal tool for porosity estimation. Fig. 12 shows the porosity and void volume distribution estimated using the different tools. The analysis represents the $Sample_{D1}$ printed from printer 2 (MEX AM process) for the first process parameters. The first process parameters correspond to 50 µm layer height, 30 mm/s printing speed, and 100 % infill density.

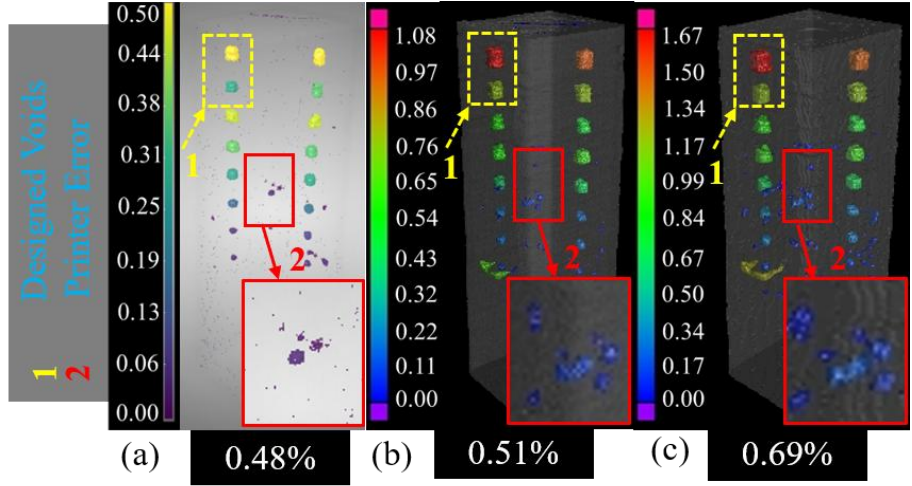

**Fig. 12** Porosity distribution estimated using different tools: **(a).** Software 1 **(b).** Software 2 **(c).** AI model

Fig. 12(a) shows the porosity distribution estimated using Software 1, 12(b) represents the analysis using Software 2, and 12(c) represents the analysis for segmented data using the AI model. The porosity estimated by Software 1, Software 2, and AI model are 0.48 %, 0.51 %, and 0.69 % respectively. It is observed that the 3D printed sample contains the designed voids that are present in the CAD model as well as the voids that exist inside the sample due to the printer error. The dashed yellow box and arrow marked as 1 represent the designed voids, while the red solid box and arrow marked as 2 are the voids created inside the sample during the MEX AM process due to the low performance of the 3D printer. The porosity present in the sample due to printer error is not accurately segmented by software 1 and 2 based on the classical image segmentation method, as shown in Fig. 12(a) and 12(b). The defects present in the sample due to printer error are segmented and visualized accurately after segmentation using the AI model, as shown in Fig. 12(c). Also, the size of the designed voids is accurately estimated by AI segmentation compared to classical segmentation methods. Therefore, AI-based segmentation is optimal tool for accurate defect detection and porosity estimation.

The quantitative value of total porosity is also estimated and analyzed for the 3D printed samples using the three tools (Software 1, Software 2, and AI model). The samples containing accurate sized designed voids and minimum defects due to printer error are better in terms of QA. Therefore, the quantitative value of total porosity for $Sample_{D1}$ printed from the different printers is estimated and plotted (Fig. 13) as a function of printer type and tool used for the analysis. The corresponding process parameters for printers 1, 2, and 3 are first (50µm layer height, 30mm/s nozzle speed, 100% infill density), first (50µm layer height, 30mm/s nozzle speed, 100% infill density), and third (60µm layer height, 30mm/s nozzle speed, 100% infill density) as listed in Table 3. These respective parameters correspond to the minimum defects present in the printed sample. The For $Sample_{D2}$, the plot is shown in Fig. SF4 of supplementary section S3.

**Table 4:** Porosity estimation using different tools

| Printer | Porosity estimated (%) using different tools. | | |
|---|---|---|---|
| | Software 1 | Software 2 | AI Model |
| Printer 1 | 2.27 | 3.01 | 3.88 |
| Printer 2 | 0.48 | 0.51 | 0.69 |
| Printer 3 | 0.89 | 0.92 | 1.13 |

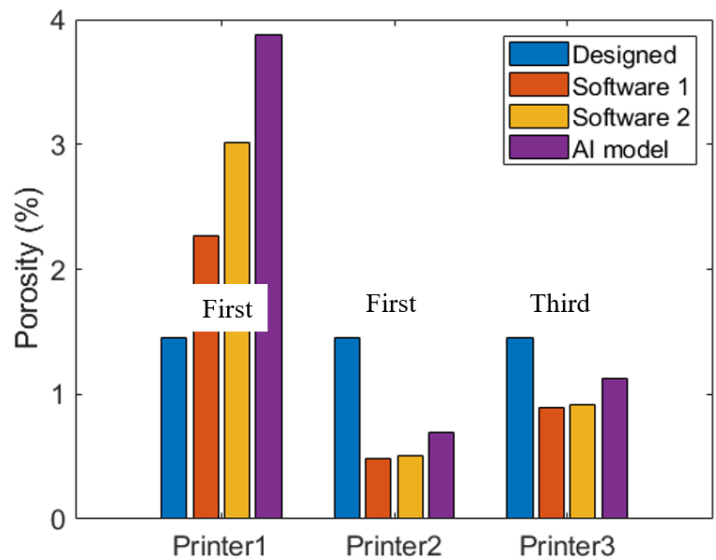

**Fig. 13** Quantitative values of porosity estimated using different tools in $Sample_{D1}$ printed from different printers.

The quantitative values of porosity estimated from the different tools are listed in Table 4. The designed porosity in $Sample_{D1}$ is ~1.45 %. It is observed from Fig. 13 and Table 4 that the porosity in the samples printed from the 3D printers (printer 1, 2, and 3), based on the MEX printing process, is either more or less. This is because other printers cannot fabricate samples with the exact resolution stated by the manufacturer or because of changes in performance over time. Therefore, the porosity analysis shows that the QA of 3D printed samples is better for the Printer 3. Fig. 13 and Table 4 also show that the porosity estimated after AI segmentation is higher than the direct porosity estimation by classical methods (Software 1 and 2). That means AI segmentation accurately detects the smaller defects printed in the sample during the MEX AM process. This also validates the accuracy of AI-based segmentation over the classical thresholding method.

### 3.3. Void distribution analysis for different printer types

The 3D volume size distribution of voids inside the samples after segmentation using the AI model 1 is analyzed using Software 2 to analyze the print quality as a function of printer type. The analysis for $Sample_{D1}$ printed from the first process parameters (50μm layer height, 30mm/s nozzle speed, 100% infill density) is shown in Fig. 14. The analysis for $Sample_{D2}$ using Software 2 is shown in Fig. SF5 of supplementary section S3. The Software 1 analysis of void size distribution for the first process parameters is shown in Fig. SF6 of supplementary section S3.

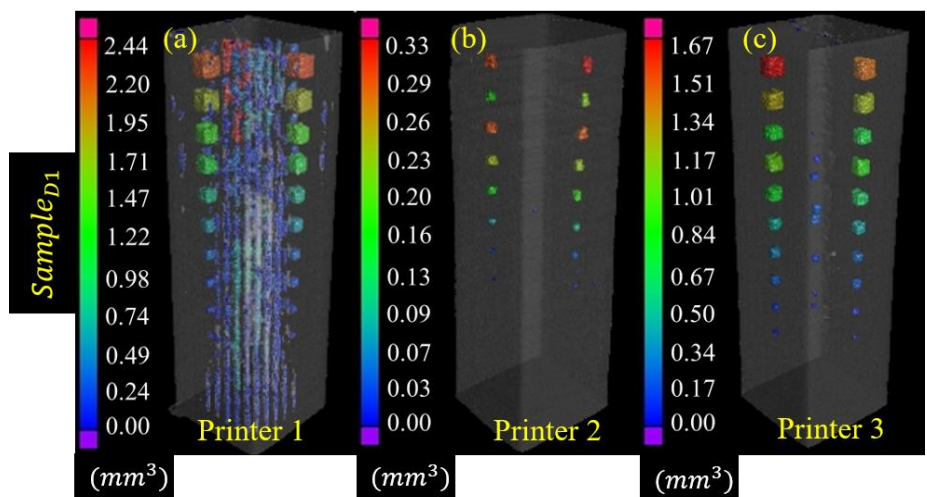

**Fig. 14** Volumetric analysis of pores analyzed using Software 2 in $Sample_{D1}$ printed from Printer 1, Printer 2, and Printer 3 at first process parameters

From Fig. 14, it is observed that voids of size 0.3mm (with 50% accuracy) and above are printed and visualized in the volumetric analysis for the samples printed from printers (1,2 and 3) based on the MEX process. The size and shape of the designed void is accurate in printer 3 (Fig. 14 (c)), followed by printer 2 (Fig. 14 (b)) and 1 (Fig. 14 (a)). Also, the internal volume of $Sample_{D1}$ printed by Printer 1 is distorted, and defects are present. This may be due to either the performance of the printer being decreased or the nozzle speed or size not being accurate for printing the voids perfectly. It is observed that the voids printed in $Sample_{D1}$ by printer 2 are of imperfect shapes. The voids printed by Printer 3 are close to the shape and size of the inserted void. The same conclusion holds for $Sample_{D2}$ shown in Fig. SF7 of supplementary data. It can also be concluded that the quality of different 3D printed samples as a function of printer type follows the order of Printer 3, Printer 2, and Printer 1. The analysis concludes that out of three MEX AM process-based printers, printer 3 is the optimal printer type for our study.

## 3.4. AI model for process control

The AI model 2 has been trained with 20 sets of 4 input variables (layer height, nozzle speed, infill density, and printer type) and 1 output variable (porosity). The printers 1, 2, and 3 are labeled as 11, 21, and 31 in training the model. The data is listed in Table Sup 5 of supplementary data section S6. The implementation of (4-4-1) layered neural network is shown in Fig. 15.

The correlation coefficient for total data is equal to 0.60538, as shown in Fig. 16. The training and validation data have a determination coefficient of 0.83 and 0.94. The higher value of R shows the accuracy of the AI model 2. This can be further improved by increasing the data set. The best validation performance obtained is 0.072797 at epoch 681. [R2 C2b:] The trained AI model 2 is then tested for a set of process parameters. The porosity predicted by the AI model 2 for the 60µm layer height, 30mm/s nozzle speed, 100 % infill density, and Printer 3 process parameters is 1.40 % which is closest to the designed amount of porosity (1.45 %). That means for this set of process parameters, the voids in the printed sample is closest to the designed model and defects due to printing process parameters is minimum. These values of AM process parameters are the optimal process parameters for our study as predicted by AI model. This concludes the successful implementation of NDT-trained (AI-estimated porosity data) AI model for process parameters control and optimization. The predicted porosity values for some of the input variables are listed in Table 5.

**Table 5:** Prediction by AI model 2

| Input Variables | | | | Output Variable | Error |
|---|---|---|---|---|---|
| Layer Height (µm) | Nozzle speed (mm/s) | Infill Density (%) | Printer Type | Porosity Predicted by AI model 2 (%) | (%) |
| 60 | 30 | 100 | Printer 2 | 1.53 | 5.51 |
| 60 | 30 | 100 | Printer 3 | 1.40 | 3.44 |
| 50 | 30 | 100 | Printer 1 | 2.36 | 4.83 |

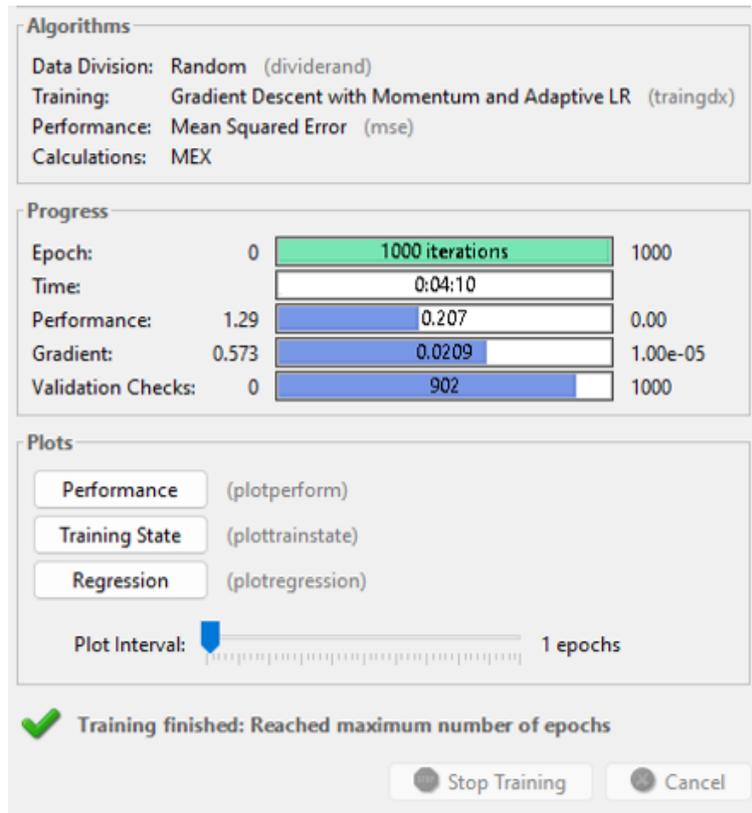

**Fig. 15** Implementation of (4-4-1) layered Artificial Neural Network for process parameters control.

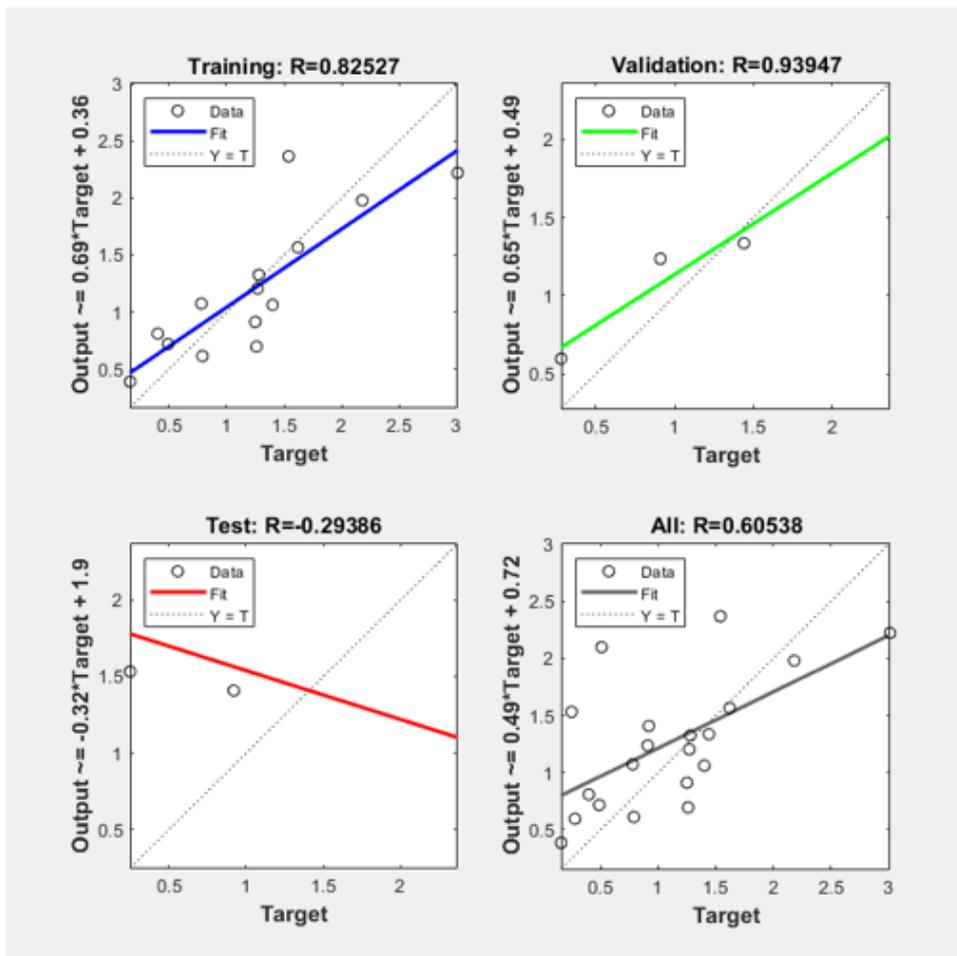

**Fig. 16** Correlation coefficient, R of data in (4-4-1) layered ANN.

### 3.5. Qualitative Analysis of Surface Roughness

The quality of the sample printed from the AM process depends upon the surface roughness present in the sample. Therefore, to confirm the results of the optimal process parameters obtained by implementing the NDT data-trained AI pipeline in sections 3.1-3.4, surface roughness analysis is performed. The qualitative analysis of surface roughness for $Sample_{D1}$ is shown in Fig. 17.

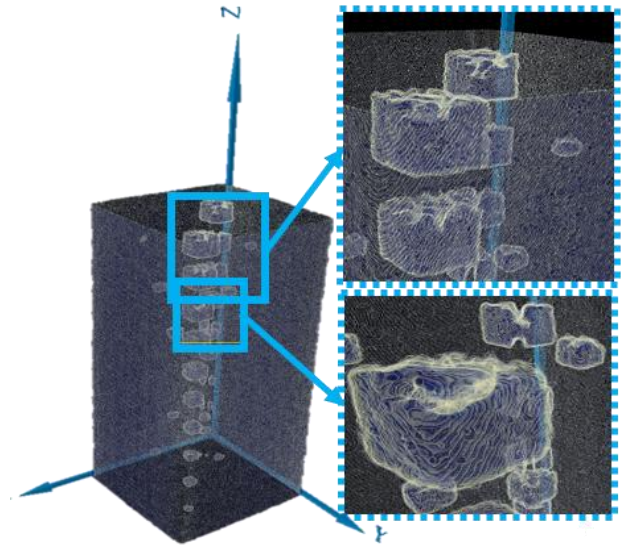

**Fig. 17** Qualitative analysis of Surface Roughness in
$Sample_{D1}$ printed by Printer 2 based on the MEX AM process
at first process parameters.

Fig. 17 illustrates the visualization of voids printed by Printer 2 based on MEX AM process in $Sample_{D1}$ for first process parameters (50µm layer height, 30mm/s nozzle speed, 100% infill density). It is observed that the designed voids are not perfectly cubical. There is distortion in the printed geometry of the void. Secondly, pores also existed in the printed sample due to an error in the nozzle. The material is not deposited entirely by the nozzle as it moves from one point to another and from one layer to another. These defects contribute to the surface roughness accumulated in the sample printed from the MEX AM process.

### 3.6. Optimal process parameters using Surface Roughness

The results of the qualitative surface roughness analysis in section 3.5 depict that the QA of the sample can also be done by estimating the surface roughness of the printed samples. Therefore, the quantitative surface roughness analysis is performed for all the 3D printed samples to find the process parameters corresponding to the minimum surface roughness and validate the results obtained from the AI-integrated pipeline in sections 3.1-3.4. The average roughness parameters $R_a$ and skewness parameters $R_{sk}$ are estimated for the 3D printed samples. The parameters are calculated separately for XY (Transverse) and XZ (Saggital) surfaces of CT slices. The box plots for minimum $R_a$ and $R_{sk}$ values for the samples printed from printers 1, 2, and 3 are shown in Fig. 18, 19, and 20, respectively. The sample sizes for printers 1, 2, and 3 are 3, 6, and 6, respectively.

The first process parameter is the optimal setting with minimum roughness parameters $R_a$ for both the samples printed from Printer 1 and the XY and XZ surfaces of the samples. The first process parameter is the minimum skewness parameter $R_{sk}$ setting for $Sample_{D1}$ and $Sample_{D2}$ printed from Printer 1 and for both XY and XZ surfaces of the samples.

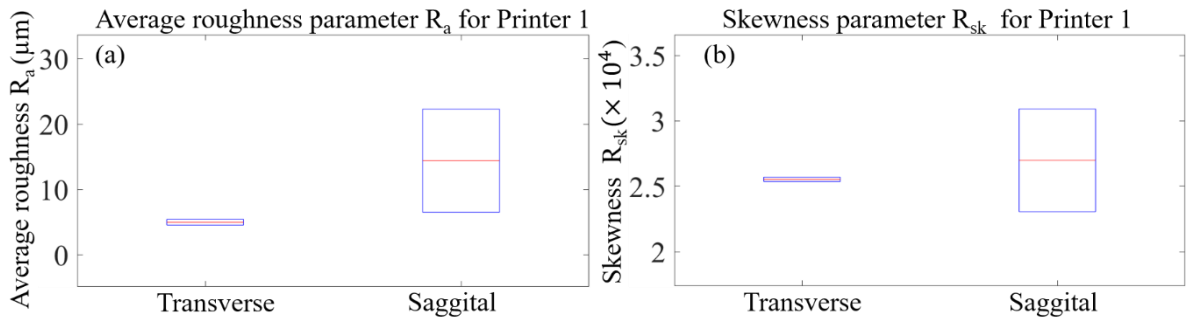

**Fig. 18** Estimation of average Roughness Parameters $R_a$ and skewness $R_{sk}$ for sample printed from printer 1.

For $Sample_{D1}$ printed from Printer 2, the minimum roughness parameter $R_a$ for the XY surface is found for the first process parameter, and for the XZ surface, it is so for the sixth process parameter. For $Sample_{D2}$ printed from Printer 2, the minimum roughness parameter $R_a$ for XY surface is found for the fifth process parameter. The first process parameter is the minimum surface roughness setting for the XZ surface. For skewness measurement, the minimum skewness parameters $R_{sk}$ for XY surface for $Sample_{D1}$ printed from Printer 2 is found at the first process parameter. The third process parameter is the minimum skewness setting for the XZ surface. For $Sample_{D2}$ printed from Printer 2, the third process parameter gives the minimum skewness parameter $R_{sk}$ for XY and XZ surfaces.

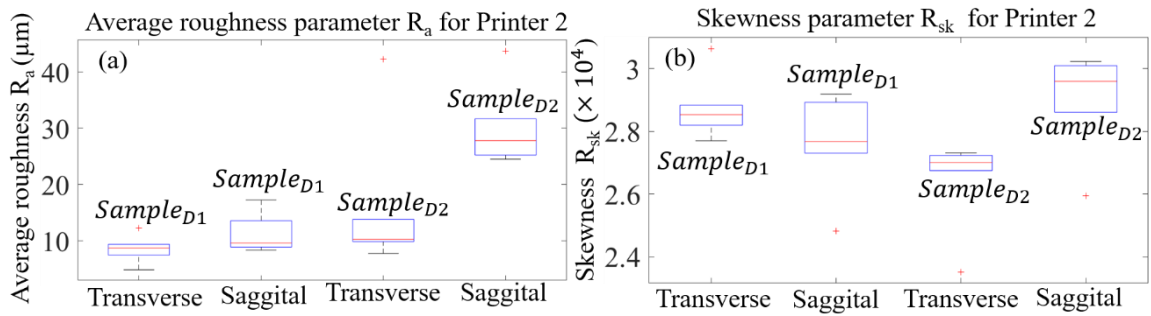

**Fig. 19** Estimation of average Roughness Parameters $R_a$ and skewness $R_{sk}$ for sample printed from printer 2.

For samples printed from Printer 3, minimum $R_a$ is found for the third process parameter for the samples for both the XY and XZ surfaces. For samples printed from Printer 3, the minimum skewness parameter $R_{sk}$ is found for the second process parameter for both the samples and XY and XZ surfaces.

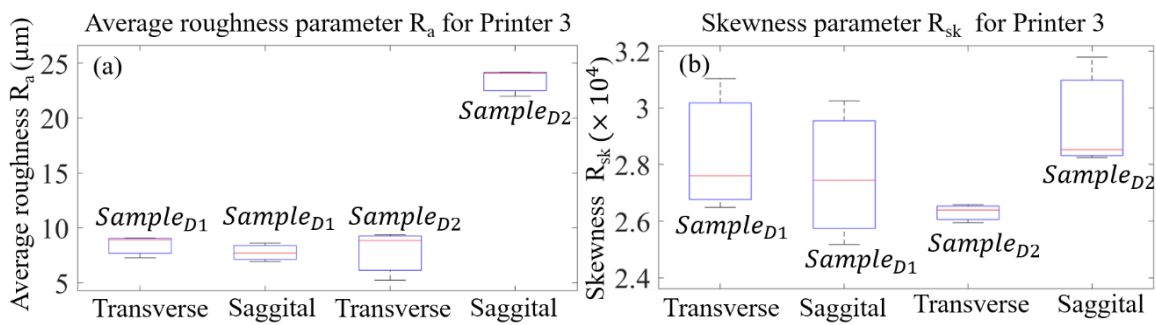

**Fig. 20** Estimation of average Roughness Parameters $R_a$ and skewness $R_{sk}$ for sample printed from printer 3.

## 3.7. Validation of proposed methodology by Mechanical testing

The compression test of different 3D printed samples with varying proportions of designed and un-designed porosity is conducted. The compressive force and stress are estimated and plotted as a function of porosity, as shown in Fig. 21(a) and 21(b), respectively. The results show that increased porosity decreases compressive force and stress value. That is, the strength of the samples decreases with the increasing porosity. This is because porous samples are less dense and, therefore, break easily. However, a slight fluctuation in the estimation is observed for the porosity value up to 0.50 %, which concludes that smaller pore size has a lesser impact on the strength of the materials/ samples. The compression test results predict that the sample with 3.01 % porosity breaks at a maximum force of 2.95kN and displacement of 6mm. The samples with a lesser amount of porosity do not break but yield. The maximum compressive force and stress value increases when porosity is decreased. [R2 C4:] The maximum compressive force (5.34 kN) and stress (83.95 MPa) value corresponds to the sample with the least porosity (0.16 %) and printed from printer 1 with first process parameters (50µm layer height, 30mm/s nozzle speed, and 100% infill density). The compressive force and stress value for the sample with porosity (1.40 %) and optimal process parameters (60µm layer height, 30mm/s nozzle speed, 100 % infill density, and Printer 3) as predicted by the proposed methodology are 2.52 kN and 70.09 MPa, respectively. These values are *in-between* the minimum and maximum limits of compressive force [1.19 5.34] kN and stress [33.23 83.95] MPa values of our study. Therefore, the optimal value of strength obtained from the mechanical testing method for the optimal process parameters predicted from the proposed methodology, validates the accuracy of the proposed methodology.

Also, the DT method fails to predict the actual strength for smaller porosity values. Therefore, the NDT method is required to accurately predict the QA of printed samples. The NDT data-trained AI method saves material wastage and is cost-effective. The samples after the compression test are shown in Fig. SF7 of supplementary section S5.

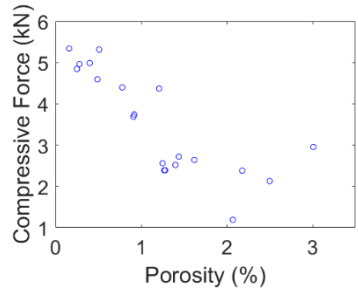

**Fig. 21(a)** Compressive Force (kN) as a function of porosity

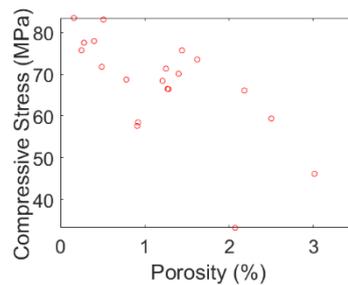

**Fig. 21(b)** Compressive Stress (MPa) as a function of porosity

## 4. Conclusion

The proposed study provides a methodology to improve the print quality of samples printed from AM process by optimizing the process parameters using NDT data trained AI models. The two AI model integrated in a series pipeline are proposed and tested for AM process parameters optimization and automation. The results obtained from the proposed methodology are verified by classical optimization method (surface roughness analysis) and mechanical testing (compression test).

The proposed methodology improves the accuracy of the method for process parameters optimization by minimizing the human intervention. Also, the data used for the training purpose is NDT data which saves material wastage because DT analysis destroys the samples. Therefore, the proposed methodology saves time by complete involvement of AI, reduces material wastage and saves energy and environment, hence, providing eco-friendly solution.

## 5. Discussion

The MEX AM process is used as a tool in the analysis due to the availability of this technology. However, the methodology can be applied for other list of process parameters and AM process. X-ray CT is also used as a tool for NDT data collection. NDT data obtained from some other technique can also be used in training the AI models. Also, presented methodology is applicable, if user wish to use surface roughness, dimensional deviation or un-designed porosity as a parameter to train the AI model and then select the optimal process parameters, accordingly. [R2 C3b:] Mutual dependency of several structural parameters (such as porosity, surface roughness, etc.,) to optimizes the process parameters can be a future investigation. The process for inclusiveness of other process parameters for any other AM process will remain the same, the results may vary depending upon the type of sample, material, process parameter, AM process type and parameter used for defect estimation.

The results of our analysis depict that AI model (U-net) is the optimal tool for porosity estimation and defect detection in the 3D printed samples. [R2 C2a:] Optimal process parameters when are set on a 3D printer, it is expected to 3D print the sample with minimum deviation from the CAD model. Printer 3 is the optimal printer for the sample printed from MEX AM process. The corresponding optimal process parameters are 60 µm layer height, 30mm/s nozzle speed, and 100 % infill density for sample design D1 printed from PLA. [R2 C5:] The compressive force and stress value for the sample with porosity (1.40 %) and optimal process parameters (60µm layer height, 30mm/s nozzle speed, 100 % infill density, and Printer 3 as predicted by the proposed methodology) are 2.52 kN and 70.09 MPa, respectively. These values are between the minimum and maximum limits of compressive force [1.19 5.34] kN and stress [33.23 83.95] MPa values of our study.

**Future work:** The methodology will be tested by including other process parameters for different AM processes.

**Acknowledgments:**

SK is thankful to the IITR Institute Assistantship. MG is thankful to the Tinkering Lab IIT Roorkee Staff.

**Statements and Declarations**:-

1. Ethics approval and consent to participate: Not required/applicable
2. Consent for publication: all authors have read the manuscript and provided their consent after significant contribution.
3. Availability of data and material: data will be provided on request.
4. Competing interests: authors declare no competing interests.
5. Funding: Not applicable.
6. Authors' contributions: SK: Writing and experiments, AD: provided access to 3D printers, MG: Funding, writing, and data processing.
7. Authors' information (optional): Provided on the title page.
8. Acknowledgments: SK acknowledges MHRD for funding.